\documentclass[letterpaper, 10 pt, conference]{ieeeconf}  % Comment this line out if you need a4paper

\IEEEoverridecommandlockouts                              % This command is only needed if 
\usepackage{graphics} % for pdf, bitmapped graphics files
\usepackage{epsfig} % for postscript graphics files
\usepackage{times} % assumes new font selection scheme installed
\usepackage{amsmath,amssymb,amsfonts} % assumes amsmath package installed

\usepackage{cite}
\usepackage{multirow, booktabs, color, soul, threeparttable}
\usepackage[algoruled]{algorithm2e}
\usepackage{placeins}
\usepackage{float}
\usepackage{graphicx}

\definecolor{hl}{rgb}{0.75,0.75,0.75}
\sethlcolor{hl}

\title{\LARGE \bf
Non-Dominated Sorting Bidirectional Differential Coevolution*
}

\author{Cicero S. R. Mendes$^{1}$, Aluízio F. R. Araújo$^{2}$, and Lucas R. C. Farias$^{3}$% <-this % stops a space
\thanks{*This work was supported by Fundação de Amparo à Ciência e Tecnologia do Estado de Pernambuco (FACEPE) under grant IBPG-0628-1.03/21.}% <-this % stops a space
\thanks{$^{1}$Cicero S. R. Mendes is an MSc student in computer science at Centro de Informatica, Universidade Federal de Pernambuco, Brazil.
        {\tt\small (csrm@cin.ufpe.br)}}%
\thanks{$^{2}$Aluízio F. R. Araújo is a full professor at Centro de Informatica, Universidade Federal de Pernambuco, Brazil. 
        {\tt\small (aluizioa@cin.ufpe.br)}}%
\thanks{$^{3}$Lucas R. C. Farias is a Ph.D. candidate in computer science at Centro de Informatica, Universidade Federal de Pernambuco, Brazil. 
        {\tt\small (lrcf@cin.ufpe.br)}}%
}

\begin{document}

\maketitle
\thispagestyle{empty}
\pagestyle{empty}

%%%%%%%%%%%%%%%%%%%%%%%%%%%%%%%%%%%%%%%%%%%%%%%%%%%%%%%%%%%%%%%%%%%%%%%%%%%%%%%%
\begin{abstract}

Constrained multiobjective optimization problems (CMOPs) are commonly found in real-world applications. CMOP is a complex problem that needs to satisfy a set of equality or inequality constraints. This paper proposes a variant of the bidirectional coevolution algorithm (BiCo) with differential evolution (DE). The novelties in the model include the DE  differential mutation and crossover operators as the main search engine and a non-dominated sorting selection scheme. Experimental results on two benchmark test suites and eight real-world CMOPs suggested that the proposed model reached better overall performance than the original model.

\end{abstract}


\section{INTRODUCTION}

Many real-world problems are defined with multiple conflicting objectives and constraints, the so-called constrained multiobjective optimization problems (CMOPs). For such problems, the aim is to find a set of solutions determining the best possible tradeoffs between objectives, the Pareto-optimal set ($\mathcal{PS}$), and its corresponding solution in the objective space, the Pareto-front ($\mathcal{PF}$). The occurrence of constraints demands that the feasibility of each candidate solution in the search space has to be considered. Therefore, the complete goal for solving CMOPs entails finding a set of feasible solutions to approximate the ($\mathcal{PF}$) with suitable levels of convergence and diversity \cite{liang_et_al_2023_survey}. A CMOP is defined as 
\begin{equation}
\begin{split}
    & \text{min } \mathbf{F(x)} = [f_1(\mathbf{x}) \text{ } f_2(\mathbf{x}) \text{ }...\text{ } f_m(\mathbf{x})]^T \in \mathbb{R}^m \\
    & \text{s.t.: } 
        \begin{array}{l}
            \begin{cases}
                g_{j}(\mathbf{x}) \leq 0, \text{  } j=1,...,p\\
                h_{j}(\mathbf{x})=0, \text{  } j=p+1,...,l  \\
                x_{k}^{min}\leq x_k \leq x_{k}^{max}, \text{  } k=1,...,n
            \end{cases}
        \end{array}
\end{split}
\label{eq:def-cmop}
\end{equation}
where every $\mathbf{x} = [x_1,...,x_n]^T\in S$ is an $n-$dimensional decision vector, $x_k$ is the $k$th decision variable, $S\subset\mathbb{R}^n$ is the decision space, $\mathbf{F}$ is the objective vector consisting of $m$ objective functions, $g_j(\mathbf{x})$ is the $j$th inequality constraint, $h_j(\mathbf{x})$ is the $(j-p)$th equality constraint, and $x_{k}^{min}$ and $x_{k}^{max}$ are the boundaries of $x_{k}$. The level of $j$th constraint violation for $\mathbf{x}$ is 
\begin{equation}
    CV_j(\mathbf{x}) =
    \begin{cases}
        \text{max}(0, g_j(\mathbf{x})), & \text{if } j \leq p, \\
        \text{max}(0, |h_j(\mathbf{x})| - \epsilon), & \text{otherwise.}
    \end{cases}
    \label{eq:def-cv}
\end{equation}
where $\epsilon$ is a small positive value (e.g $\epsilon=10^{-4}$) to relax the equality constraints.
Hence, the overall constraint violation is  
\begin{equation}
    CV(\mathbf{x}) = \sum_{j=1}^{l}CV_j(\mathbf{x}).
    \label{eq:all-cv}
\end{equation}
where $\mathbf{x}$ is a feasible solution if $CV(\mathbf{x)}=0$; The space of feasible solutions is defined as $\mathbb{F} = \{\forall\mathbf{x}\in \mathbb{R}^n \mid CV(\mathbf{x})=0 \}$. 

CMOPs are complex optimization problems for which evolutionary algorithms (EAs) need to handle the constraints featuring the powerful capability of search to reach the solutions. Thus, the researchers often combine an EA with a constraint-handling technique (CHT). The former embodies a search engine, whereas the latter aims to guide the search process by influencing the selection of candidate solutions. However, such an EA must face many challenges due to various infeasible solutions in the search \cite{zhan2022survey}.

% #####################################################
% #####             lit review              ###########
% #####################################################

Since differential evolution (DE) was proposed by Storn and Price \cite{storn_and_price_1997}, it has been used for global optimization in many problems. DE has several favorable features such as ease of use, simplicity, efficiency, and robustness \cite{opara_and_arabas_2019}. Therefore, several DE variants with various CHTs to deal with CMOPs were proposed.

Yu \textit{et al.} \cite{yu_et_al_2021} proposed a two-phase multi-objective DE, named PDTP-MDE, in which the evolutionary process is divided into two sequential phases. In the first phase, the aim is to achieve a balance between convergence and diversity; meanwhile, feasibility is used as an assistant criterion. In the second phase, the diversity is maintained along with the use of feasible and promising infeasible solutions according to the population evolution. Feasible solutions are stored and updated in an archive to form the Pareto optimal solutions.

Yang \textit{et al.}\cite{yang_et_al_2019_mode_sae} presented a multi-objective differential evolution algorithm (MODE-SaE) that integrates an improved epsilon CHT mechanism, a maintenance strategy of elite feasible solutions archive, and self-switching parameters of search engine. The improved epsilon CHT mechanism self-adaptatively adjusts the epsilon level according to the constraint violation values of the candidate solutions. The feasible solutions are saved to the external archive. Subsequently, the same authors proposed a multi-objective differential evolution algorithm based on domination and a CHT switching mechanism named MODE-CHS \cite{yang_et_al_2021_mobj_de}. In the switching mechanism, if there is no feasible solution in the population, the evolution process runs with a CHT; otherwise, it runs without a CHT. The feasible solutions are stored in an archive that is evolved jointly with the population. 

Wang \textit{et al.}\cite{wang_et_al_2019_ccmode}  proposed a cooperative DE framework for CMOPs, called CCMODE, in which the algorithm maintains $M$ constrained single-objective subpopulations and a population archive for a $M$-objective optimization problem. Those $(M+1)$-populations co-evolve to optimize CMOPs. Wang \textit{et al.}\cite{wang_et_al_2019_c2ode} presented an extension of CoDE \cite{wamg_et_al_2011}, called $C^2oDE$. It employs three complementary trial vector generation strategies to handle convergence and diversity. They also proposed a CHT based on feasibility rules and the epsilon constraint method.

Among the CMOEAs with different CHTs for handling CMOPs, we can find several studies based on DE \cite{liang_et_al_2023_survey}. The study by Liu and Wang \cite{liu_and_wang_2019} suggested that DE was more effective in finding feasible solutions in three CMOEAs. Aiming at extending the DE capability and based on the bidirectional coevolution algorithm (BiCo) \cite{liu_et_al_2022}, a recent promising model to solve CMOPs, this paper proposes the non-dominated sorting bidirectional differential coevolution (NSBiDiCo). This model employed the DE differential mutation and crossover operators, and a selection scheme inspired by \cite{angira_2005_nsde, deb_et_al_2002}. NSBiDiCo uses the constraint dominance principle (CDP) and a crowding distance schema for choosing survivors between target and trial vectors. The experimental results in two benchmark test suites and eight real-world problems suggested the efficacy of NSBiDiCo.

% #####################################################
% #####           end lit review            ###########
% #####################################################

In this paper, Section II describes in detail the proposed algorithm, NSBiDiCo. Section III presents the experimental setup, Section IV discusses the experimental results, and the conclusion and future work are given in Section V.

\SetKwComment{Comment}{ /* }{ */}

\section{NSBiDiCo}
\begin{figure*}[htb]
    \centering
    \includegraphics[width=1\linewidth, keepaspectratio]{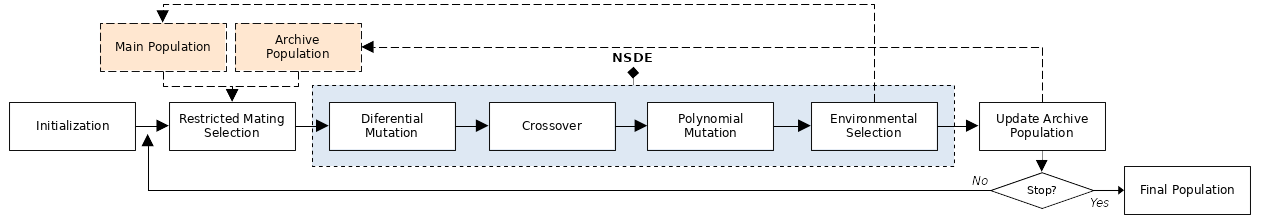}
    \caption{The flowchart of the proposed NSBiDiCo algorithm.}
    \label{fig:NSBiDiCo_flow_chart}
\end{figure*}
This paper proposes a modification of BiCo \cite{liu_et_al_2022}, an algorithm to solve CMOPs via bidirectional coevolution. Fig. \ref{fig:NSBiDiCo_flow_chart} depicts the main framework of NSBiDiCo. The detailed procedure is given in Algorithm \ref{alg:NSBiDiCo}. The first step consists of initializing the main population $\mathcal{P}_0=\mathbf\{x_1, x_2,..., x_N\}$  with random samples in the search space, the archive population $\mathcal{A}_0$ is made void. Then, a restricted mating selection chooses candidate solutions for trial solution generation. Subsequently, the Non-dominated Sorting Differential Evolution (NSDE) \cite{angira_2005_nsde} is performed, where the differential mutation and crossover operators produce the trial solutions $\mathcal{Q}_t$. It is worth mentioning that after this step, a polynomial mutation procedure is also performed \cite{li_and_zhang_2009}. The selection for the new population $\mathcal{P}_{t+1}$ is based on the constraint dominance principle. Therefore, we use a non-dominated sorting procedure to order all target and trial vectors \cite{deb_et_al_2002}. To update the archive population $\mathcal{A}_{t+1}$, a non-dominated sorting procedure and an angle-based selection are executed as it occurs for BiCo. All these steps are performed in a loop until a stop criterion is reached.

\begin{algorithm}
    {\scriptsize
        \caption{NSBiDiCo Framework}
        \label{alg:NSBiDiCo}
        \SetKwInOut{Input}{Input}\SetKwInOut{Output}{Output}
        \Input{A CMOP, population size $N$, recombination probability $Cr$, scaling factor $F$, mutation probability $p_m$, and distribution index $\eta_m$}
        \Output{$\mathcal{P}_{t+1}$}
        Initialize $t=0$, $\mathcal{P}_{0}\leftarrow\{\mathbf{x}_1, \mathbf{x}_2,...,\mathbf{x}_N\}$, and $\mathcal{A}_{0}\leftarrow\varnothing$;
    
        \While{$\text{the stop criterion is not met}$}{
            Select mating parents $\overline{\mathcal{P}}$ from $\mathcal{P}_t$ and $\mathcal{A}_t$ (Alg. \ref{alg:NSBiDiCo_rmp})\;
            Execute DE variation operators to produce trial solutions $\mathcal{Q}_t$ from $\overline{\mathcal{P}}$\;
            
            Update $\mathcal{P}_{t+1}$ from $\mathcal{P}_{t}$ and $\mathcal{Q}_{t}$ (Alg. \ref{alg:NSBiDiCo_ump})\;
            
            Select the non-dominated infeasible solutions $\mathcal{V}_t$ from $\mathcal{P}_{t}\cup\mathcal{A}_{t}\cup\mathcal{Q}_t$\;
            Run the angle-based selection scheme to update $\mathcal{A}_{t+1}$ from $\mathcal{V}_t$ (Alg. \ref{alg:NSBiDiCo_uap})\;
            $t=t+1$\;
        }
    }
\end{algorithm}

NSBiDiCo runs a selection procedure (Algorithm \ref{alg:NSBiDiCo_ump}) to update the main population $\mathcal{P}$. Initially, a non-dominated sorting is run on the set $\mathcal{U}_t = \mathcal{P}_t\cup\mathcal{Q}_t$ and the entire $\mathcal{U}_t$ are ranked to build different fronts (i.e, $\mathcal{F}_1, \mathcal{F}_2,...,\mathcal{F}_n$) according to their non-domination levels. Then, $N$ solutions of $\mathcal{U}_t$ are selected to the next population $\mathcal{P}_{t+1}$. If the size of the front $\mid\mathcal{F}_i\mid$ plus $\mid\mathcal{P}_{t+1}\mid$ exceeds $N$, the crowding distance is used to sort the solutions in descending order and $N - \mid\mathcal{P}_{t+1}\mid$ solutions are selected from $\mathcal{F}_i$. On the other hand, $\mathcal{F}_i$ is directly assigned to the next population, and the iteration goes to the next front level $\mathcal{F}_{i+1}$ until $\mid\mathcal{P}_{t+1}\mid$ reach size $N$. This is an elitist mechanism \cite{deb_et_al_2002}.

\begin{algorithm}
    {\scriptsize
        \caption{Environmental Selection}
        \label{alg:NSBiDiCo_ump}
        
        \SetKwInOut{Input}{Input}\SetKwInOut{Output}{Output}
    
        \Input{$\mathcal{P}_t$ and $\mathcal{Q}_t$}
        \Output{$\mathcal{P}_{t+1}$}
        $\mathcal{U}_t = \mathcal{P}_t \cup \mathcal{Q}_t$\;
        
        $\mathcal{F} \leftarrow$  run a fast non-dominated sorting procedure in $\mathcal{U}_t$\;
        $\mathcal{P}_{t+1} \leftarrow \varnothing$ and $i\leftarrow1$\;
        
        \While{$\mid\mathcal{P}_{t+1}\mid + \mid\mathcal{F}_i\mid \leq N$}{
            $\mathcal{C}_i \leftarrow $ crowding distance of $\mathcal{F}_i$\;
            $\mathcal{P}_{t+1} \leftarrow \mathcal{P}_{t+1} \cup \mathcal{F}_i$\;
            $i\leftarrow i+1$\;
        }
        $\mathcal{F}_i \leftarrow$ sort according to $\mathcal{C}_i$ values in descending order\;
        $\mathcal{P}_{t+1} \leftarrow \mathcal{P}_{t+1} \cup \mathcal{F}_i[1:(N - \mid\mathcal{P}_{t+1}\mid)\big]$\;
    }
\end{algorithm}

The population in the archive used by BiCo is the primary force toward the $\mathcal{PF}$ from the infeasible region in the search space. The Algorithm \ref{alg:NSBiDiCo_uap} describes the update of the archive in two steps: determining non-dominated infeasible solutions and an angle-based selection scheme.

\subsection{Determination of the non-dominated Infeasible Solutions}

Such solutions are found following the steps:
\begin{enumerate}
    \item Take $CV$ in \eqref{eq:all-cv} as an additional objective function and transform the original CMOP in \eqref{eq:def-cmop} into an unconstrained $(m+1)$-objective MOP
    \begin{equation}
        \label{eq:def-mop-cv}
        \text{min   } \mathbf{F(x)} = [f_1(\mathbf{x}), f_2(\mathbf{x}), ..., f_m(\mathbf{x}), CV(\mathbf{x})]^T.
    \end{equation}
    \item Run a non-dominating sorting procedure to sort the non-dominated solutions for the union of the main population $\mathcal{P}_t$, archive population $\mathcal{A}_t$, and trial solutions $\mathcal{Q}_t$ according to \eqref{eq:def-mop-cv}.
    \item Select the infeasible non-dominated solutions to update $\mathcal{V}_t$.
\end{enumerate}
\begin{algorithm}
    {\scriptsize
        \caption{Updating of the Archive Population}
        \label{alg:NSBiDiCo_uap}
        
        \SetKwInOut{Input}{Input}\SetKwInOut{Output}{Output}
    
        \Input{$\mathcal{P}_t, \mathcal{A}_t$ and $\mathcal{Q}_t$}
        \Output{$\mathcal{A}_{t+1}$}
        $\mathcal{U}_t \leftarrow \mathcal{P}_t \cup\mathcal{A}_t\cup\mathcal{Q}_t$\;
        Select the infeasible non-dominated solutions $\mathcal{V}_t$\ in $\mathcal{U}_t$ by regarding Eq. \eqref{eq:def-mop-cv}\;
           % Find the nondominated solutions in $\mathcal{U}_t$ by regarding Eq. \eqref{eq:def-mop-cv}, and select the infeasible nondominated solutions $\mathcal{V}_t$\;
        \While{$\mid\mid\mathcal{V}_t\mid\mid > N$}{
            Find the two solutions $\mathbf{u_i}$ and $\mathbf{u_j}$ with the smallest vector angle ($\Theta_{\mathbf{u}_i, \mathbf{u}_j}$) in $\mathcal{V}_t$\;
            \eIf{$CV(\mathbf{u_j}) < CV(\mathbf{u_i})$}{
                $\mathcal{V}_t \leftarrow \mathcal{V}_t\setminus\mathbf{u_i}$ \Comment*[r]{Delete $\mathbf{u_i}$}
            }{
                $\mathcal{V}_t \leftarrow \mathcal{V}_t\setminus\mathbf{u_j}$ \Comment*[r]{Delete $\mathbf{u_j}$}
            }
        }
        $\mathcal{A}_{t+1} \leftarrow \mathcal{V}_t$\;
    }
\end{algorithm}

\subsection{Angle-based Selection}

If the cardinality of $\mathcal{V}_t$ is larger than the archive size $N$, then the angle-based selection scheme is performed to discard exceeding solutions. 

The angle-based selection scheme deals with the angle between two solutions in the normalized objective space (without considering the constraints). To compute the angle vector, we estimate both the ideal point $\mathbf{Z}^{min} = [z^{min}_{1}, z^{min}_{2},...,z^{min}_{m}]^T$ and the nadir point $\mathbf{Z}^{max} = [z^{max}_{1}, z^{max}_{2},...,z^{max}_{m}]$, where $\mathbf{z}^{min}_{i}$ and $\mathbf{z}^{max}_{i}$ denote the minimum and maximum values of the $i$th objective for all solutions in $\mathcal{V}_t$. After that, the objective vector $\mathbf{F}(\mathbf{v}_j)$ is normalized as $\mathbf{F^{'}}(\mathbf{v}_j) = [f_1(\mathbf{v}_j), f_2(\mathbf{v}_j),..., f_m(\mathbf{v}_j)]^T$ according to
\begin{equation}
    f_{i}^{'}(\mathbf{v}_j) = \frac{z^{max}_{i} - f_i(\mathbf{v}_j)}{z^{max}_{i} - z^{min}_{i}}, i = 1,2,...,m.
\end{equation}
Then, the vector angle $\theta$ between two solutions $\mathbf{v}_j$ and $\mathbf{v}_k$ is
\begin{equation}
    \theta_{\mathbf{v}_j,\mathbf{v}_k} = \text{arccos } \Bigg|\frac{ \mathbf{F}^{'}(\mathbf{v}_j) \bullet \mathbf{F}^{'}(\mathbf{v}_k)} { \parallel\mathbf{F}^{'}(\mathbf{v}_j)\parallel \cdot \parallel\mathbf{F}^{'}(\mathbf{v}_k)\parallel}\Bigg|
\end{equation}
where $\mathbf{F^{'}}(\mathbf{v}_j) \bullet \mathbf{F^{'}}(\mathbf{v}_k)$ is the product between $\mathbf{F^{'}}(\mathbf{v}_j)$ and $\mathbf{F^{'}}(\mathbf{v}_k)$, and $\parallel\cdot\parallel$ is the norm of the vector. As pointed out in \cite{liu_et_al_2022}, the vector angle between solutions estimates the difference between the considered vectors.

To maintain diversity, poor solutions are deleted one by one from the archive population. To this end, two solutions in $\mathcal{V}_t$ with the minimum vector angle between them are identified. Since solutions with smaller angles between them search in the same direction, the deletion procedure considers the total of constraint violation $CV$ to make a choice. Therefore, the solution with the higher $CV$ is deleted from $\mathcal{A}_t$. These steps will continue while the size of $\mathcal{V}_t$ is greater than $N$.
\begin{algorithm}
    {\scriptsize
        \caption{Restricted Mating Pool}
        \label{alg:NSBiDiCo_rmp}        
        \SetKwInOut{Input}{Input}\SetKwInOut{Output}{Output}
        \Input{$\mathcal{P}_t$ and $\mathcal{A}_t$}
        \Output{Mating parents $\mathbf{p_1}$ and $\mathbf{p_2}$}
    
        \eIf{$\mid\mid\mathcal{A}_t\mid\mid < N$}{
            Randomly select two solutions $\mathbf{p_1}$ and $\mathbf{p_2}$ from $\mathcal{P}_t \cup\mathcal{A}_t$\;        
        }{
            Randomly select two solutions $\mathbf{x_1}$ and $\mathbf{a_1}$ from $\mathcal{P}_t\cup\mathcal{A}_t$, respectively\;
    
            \eIf{$CV(\mathbf{x_1}) \leq CV(\mathbf{a_1})$}{
                $\mathbf{p_1} \leftarrow \mathbf{x_1}$\;
            }{
                $\mathbf{p_1} \leftarrow \mathbf{a_1}$\;
            }
            Randomly select two solutions $\mathbf{x_2}$ and $\mathbf{a_2}$ from $\mathcal{P}_t\cup\mathcal{A}_t$, respectively\;
    
            \eIf{$AD(\mathbf{x_2}) \leq AD(\mathbf{a_2})$}{
                $\mathbf{p_2} \leftarrow \mathbf{x_2}$\;
            }{
                $\mathbf{p_2} \leftarrow \mathbf{x_2}$\;
            }
        }
    }
\end{algorithm}
The interaction between the main population $\mathcal{P}$ and the archive population $\mathcal{A}$ as a mechanism to share information is performed by the restricted mating selection proposed for BiCo algorithm. The detailed procedure is depicted in the Algorithm \ref{alg:NSBiDiCo_rmp}. In general, the following two conditions hold:
\begin{enumerate}
    \item If the size of the archive population is less than $N$, the mating parents $\mathbf{p_1}$ and $\mathbf{p}_2$ are randomly selected from the union of $\mathcal{P}_t$ and $\mathcal{A}_t$.
    \item Otherwise, $\mathbf{p}_1$ and $\mathbf{p}_2$ are selected based on $CV$ and angle-diversity value (i.e., $AD$), respectively. In this step, to select the first solution $\mathbf{p}_1$, two solutions $\mathbf{x}_1$ and $\mathbf{a}_1$ are selected from $\mathcal{P}_t$ and $\mathcal{A}_t$, respectively. Afterward, their $CV$ values are compared, and the one with the smaller $CV$ is selected. Next, to select $\mathbf{p}_2$, two solutions $\mathbf{x}_2$ and $\mathbf{a}_2$ are selected from $\mathcal{P}_t$ and $\mathcal{A}_t$, respectively; Then, their $AD$ values are compared and the one with the larger $AD$ is chosen.    
\end{enumerate}

To compute $AD(\mathbf{x})$ for $j$th solution $\mathbf{x}_j$ in $\mathcal{P}_t$ the following steps are executed.
\begin{enumerate}
    \item Find the ideal point $\overline{Z}^{min} = [\overline{z}^{min}_{1}, \overline{z}^{min}_{2},...,\overline{z}^{min}_{m}]^T$ and estimate the nadir point $\overline{Z}^{max} = [\overline{z}^{max}_{1}, \overline{z}^{max}_{2},...,\overline{z}^{max}_{m}]^T$, where $\overline{z}^{min}_{i}$ and $\overline{z}^{max}_{i}$ denote the minimum and maximum values of the $i$th objective for all solutions in joint population $\mathcal{P}_t$ and $\mathcal{A}_t$, respectively.
    \item Normalize the $j$th solution's objective vector $\mathbf{F}(\mathbf{x}_j)$ as $\mathbf{F}^{*}(\mathbf{x}_j) = [f^{*}_{1}(\mathbf{x}_j), f^{*}_{2}(\mathbf{x}_j),...,f^{*}_{m}(\mathbf{x}_j)]^T$ according to
    \begin{equation}
        f_{i}^{*}(\mathbf{x}_j) = \frac{f_i(\mathbf{x}_j) - \overline{z}^{min}_{i}}{\overline{z}^{max}_{i} - \overline{z}^{min}_{i}}, i = 1,2,...,m.
    \end{equation}
    \item Compute the vector angles between $\mathbf{x}_j$ and the other solution in $\mathcal{P}_t$ according to
    \begin{equation}
        \begin{split}
            & \theta'_{\mathbf{x}_j, \mathbf{x}_k} = \text{arccos } \Bigg|\frac{ \mathbf{F}^{*}(\mathbf{x}_j) \bullet \mathbf{F}^{*}(\mathbf{x}_k)}{\parallel\mathbf{F}^{*}(\mathbf{x}_j)\parallel \cdot \parallel\mathbf{F}^{*}(\mathbf{x}_k)\parallel}\Bigg| \\ 
            &\mathbf{x}_k \in \mathcal{P}_t \cap \mathbf{x}_k \neq \mathbf{x}_j
        \end{split}
    \end{equation}
    \item Assign $AD(\mathbf{x}_j)$ as the $k$th minimum value in the set of $\{\theta'_{\mathbf{x}_j, \mathbf{x}_k}\in\mathcal{P}_t\cap\mathbf{x}_k \neq \mathbf{x}_j\}$ where $k$ is set as $\sqrt{N}$.
\end{enumerate}
A larger value of $AD$ for $\mathbf{x}_j$ is desired. The same procedure calculates the angle-based diversity of $AD(\mathbf{a}_j)\in\mathcal{A}_t$. 

After the formation of the pool of candidate solutions $\overline{\mathcal{P}}_t$, the DE differential mutation and crossover operators generate a new set of trial solutions $\mathcal{Q}_t$. We used a $DE/current/1/bin$ defined as:
\begin{equation}
    \mathbf{v}_i = \mathbf{x}_{current} + F \times(\mathbf{x}_{r1} - \mathbf{x}_{r2})
\end{equation}
where $\mathbf{x}_{current}$ denotes the candidate solution selected as the base vector from $\overline{\mathcal{P}}$; $\mathbf{x}_{r_1}$ and $\mathbf{x}_{r_2}$ are vectors randomly selected to construct the mutant vector $\mathbf{v}_i$, where $current \neq r_1 \neq r_2$; \{$r_1, r_2\} \in [1, N]$ and $F$ is the scaling factor with value in the range $[0,1]$. 

The binomial crossover strategy follows the mutation. The crossover operator generates a set of trial vectors $\mathcal{U}_t$ where each $\mathbf{u}_i \in \mathcal{U}_t$ is produced by using the target vector $\mathbf{x}_{current}$ and mutant vector $\mathbf{v}_i$ as follows.
\begin{equation}
        u_{i,j} = \begin{cases}
        v_{i,j}, \text{    if }(rand < CR),\\
        x_{current,j}, \text{    otherwise.}
    \end{cases},
    j=1,2,...,D.
\end{equation}
where $rand$ stands for a random number sampled in the range $[0,1]$; $CR$ is a real number between 0 and 1; $D$ is the dimension of the solution; and $u_{i,j}$ is the $j$th element of the $i$th new trial vector. After that, a polynomial mutation \cite{deb1996combined} runs to perturb the solutions in $\mathcal{U}_t$ with mutation probability $p_m$. This procedure generates a mutant trial vector. For a given solution $\mathbf{u}_i \in \mathcal{U}_t$: 
\begin{equation}
    \mathbf{t}_i = \mathbf{u}_i + (\mathbf{u}_{UB,i} - \mathbf{u}_{LB,i})\times\delta,
\end{equation}
where $UB$ and $LB$ are the upper and lower bound of each component of the $i$th decision variable; The probability distribution of $\delta$ is $p(\delta) = 0.5(\eta_{m} + 1)(1 - \mid\delta\mid)^{\eta_{m}}$ and $\eta_{m}$ is the distribution index which determines the shape of the distribution. $\delta$ can be computed as follows:
\begin{equation}
    \delta(\rho) = \begin{cases}
        (2\rho)^{\frac{1}{(\eta_{m}+1)}} - 1, & \text{if } \rho \leq 0.5\\
        1 - \big[2 (1 - \rho)^{\frac{1}{(\eta_m + 1)}}\big], & \text{otherwise.}
    \end{cases}
\end{equation}
where $\rho$ is a uniform random number between $0$ and $1$.

\section{EXPERIMENTAL SETUP}

\subsection{Test Instances}

The performance of NSBiCo against BiCo is compared for two sets of test functions, LIRCMOP and DOC, and in eight real-world problems. Tables \ref{table:igd_hv_results} and \ref{tab:rwmops} present the number of objectives $M$ and decision variables $D$ of each problem.

\subsection{Parameter Settings}
\begin{enumerate}
    \item \textit{Population Size:} The size of population was set to $100$.
    \item \textit{Parameter Settings for DE:} For the benchmark test suites $CR$ and $F$ were set to their default values, $1$ and $0.5$, respectively; For the real-world problems, a Latin Hypercube Sampling with $50$ samples for $CR$ and $F$ with values between $[0,1]$ were performed. Then, the mean of the three best parameters of $CR = \{0.49, 0.4, 0.47\}$ and $F = \{0.86, 0.65, 0.59\}$ were computed and the $F$ and $CR$ were set to $0.45$ and $0.7$, respectively. Finally, for polynomial mutation, $\eta_m = 20$ and $p_m = 1/D$ were set for distribution index and mutation probability, respectively.
    
    \item \textit{Number of Independent Runs and Termination Criterion:} The experiments were independently run 30 times on each test function. For LIRCMOP, the maximum of function evaluations (FEs) was set to $300000$, while for the DOC, the maximum number of FEs was set to $200000$. Specifically, for real-world problems defined in \cite{kumar_2021}, the number of FEs was set as follows: $20000$ for vibrating platform design, welded beam design, disc brake design, Haverly’s pooling problem and reactor network design; $26250$ for car side impact design and heat exchanger network design; and, finally, $53000$ for water resources management.    
    \item For the BiCo algorithm, the parameters were kept identical to their original paper. 
\end{enumerate}

All experiments were conducted in the PlatEMO platform \cite{tian_et_al_2017} and run under Pop!\_OS 22.04 LTS on an AMD Ryzen 7 5800h with 16Gb of RAM.
 
\subsection{Performance Metrics}

The hypervolume (HV) and the inverted generational distance (IGD) have been chosen as performance metrics. The metrics were formulated according to the PlatEMO platform. It is worth mentioning that the two metrics only consider the feasible solutions of the last generation. When an algorithm finds no feasible solution over the 30 runs, the value attributed to the metric is $NaN$. The Wilcoxon rank sum test was chosen with a significance level of $0.05$ to assess the statistical differences between performances. The symbols $``+"$, $``-"$, and $``="$ indicate that the results reached by NSBiDiCo are, respectively, better, worse, or statistically equivalent to those obtained by BiCo. The HV is the only metric used for real-world problems.

\section{EXPERIMENTAL RESULTS}

This section presents the results for two test suites and six real-world problems.

\subsection{Comparison for DOC Test Suite}

The DOC \cite{liu_and_wang_2019} test suite has nine problems that simultaneously consider the objective and decision constraints and equality and inequality constraints. Wang and Liu \cite{liu_and_wang_2019} argue that such characteristics have the potential to simulate actual CMOPs. The Pareto fronts in DOC are continuous, disconnected, convex, concave, linear, mixed, degenerate, or multimodal. Moreover, the feasible regions in the decision space can be nonlinear, very small, or multimodal.

Table \ref{table:igd_hv_results} shows the performance of NSBiDiCo for DOC instances. NSBiDiCo performs best in both HV and IGD for DOC1, with concave and continuous $\mathcal{PF}$ with a nonlinear feasible region. For DOC2, with convex and disconnected $\mathcal{PF}$ plus a tiny and nonlinear feasible region, NSBiDiCo reached better performance than BiCo which did not find any feasible solution. DOC3 has concave, disconnected, and multimodal $\mathcal{PF}$ with very small, nonlinear, and multimodal search space. For this instance, NSBiDiCo did better in IGD and were statistically similar in HV. For DOC4 and DOC8, which have $\mathcal{PF}$s linear and disconnected with feasible region tiny and nonlinear, NSBiDiCo has the best performance in both HV and IGD. For DOC5, which has disconnected and multimodal $\mathcal{PF}$ with tiny and nonlinear feasible space, NSBiDiCo reached better HV and IGD than BiCo, which could not find feasible solutions. DOC6 and DOC7 have mixed and multimodal $\mathcal{PF}$s but a very small and nonlinear feasible space versus a very small and multimodal feasible space, respectively. In both instances, NSBiDiCo performed better than BiCo in both metrics. As for DOC9, which has degenerate and multimodal $\mathcal{PF}$ with very small, nonlinear, and multimodal feasible space, NSBiDiCo and BiCo performed poorly despite BiCo reaching a better IGD value.

The NSBiDiCo achieved the best overall performance on the DOC suite for both HV and IGD metrics, suggesting efficacy for CMOPs in different search spaces and Pareto fronts.

\subsection{Comparison for LIRCMOP Test Suite}

The 14 test problems on LIRCMOP \cite{fan_et_al_2020} suite are quite challenging since each of them has a number of large infeasible regions, and the feasible regions are relatively small. Some of the Pareto fronts are obstructed by such infeasible regions. Also, some of the Pareto fronts present few disjoint segments or sparse dots obstructed by large infeasible regions, and for some Pareto fronts, the feasible regions are just curves.  

Table \ref{table:igd_hv_results} suggests that NSBiDiCo has achieved better performance in IGD and HV metrics for the problems LIRCMOP1-LIRCMOP4 characterized by a curve as feasible region and with their unconstrained Pareto fronts ($\mathcal{UPF}$) far from their constrained Pareto fronts ($\mathcal{CPF}$). For LIRCMOP5 and LIRCMOP6 with large infeasible regions in front of the $\mathcal{CPF}$, NSBiDiCo reached better performance  than BiCo in HV and IGD except for IGD for LIRCMOP6 where they were statistically even. For LIRCMOP7, LIRCMOP8, LIRCMOP11, and LIRCMOP12, their $\mathcal{CPF}$s are located on their constraint boundaries. For LIRCMOP7, NSBiDiCo have similar performance in terms of both IGD and HV metrics; for LIRCMOP8 NSBiDiCo has similar performance in HV and worse in IGD; for LIRCMOP11 and LIRCMOP12, NSBiDiCo has better performance in both HV and IGD. For LIRCMOP9 and LIRCMOP10, their $\mathcal{CPF}$s are parts of their $\mathcal{UPF}$; hence, preserving the diversity of the population is the main challenge. For such problems, NSBiDiCo performed better in both HV and IGD metrics. NSBiDiCo performed better than BiCo in both HV and IGD for LIRCMOP13 and LIRCMOP14, cases with three objectives, and the $\mathcal{CPF}$ is the same as its $\mathcal{UPF}$. LIRCMOP13 and LIRCMOP14 are cases with three objectives. For LIRCMOP13, where its $\mathcal{UPF}$ are the same $\mathcal{CPF}$, NSBiDiCo has achieved better performance than BiCo in both HV and IGD. For LIRCMOP14, that have $\mathcal{UPF}$ located on the constraint boundaries, NSBiDiCo performed better in both HV and IGD. 
The results suggest that NSBiDiCo can deal with different shapes of $\mathcal{UPF}$ and $\mathcal{CPF}$ along with large infeasible regions in the search space.

\subsection{Comparison on Real-World problems}

NSBiDiCo was also competitive in real-world problems. Table \ref{tab:rwmops} shows that NSBiDiCo outperformed BiCo in $5$ out of $8$ cases, vibrating platform design, welded beam design, Haverly’s pooling problem, reactor network design, and heat exchanger network design. Only in $2$ cases, BiCo did better, on problems of car side impact design and water resources management. For disc brake design NSBiDiCo have a statistically equivalent performance to BiCo. 

\begin{table*}[!ht]
    \renewcommand{\arraystretch}{1.2}
    \centering
    \caption{Mean And Standard Deviation For Performance Comparisons Between NSBiDiCo And BiCo For DOC And LIRCMOP Test Suites.}

    \centering\scriptsize

    \begin{tabular}{clc cc | cc}
        \toprule
        \multirow{2}{*}{Problem} & 
        \multirow{2}{*}{$M$} & 
        \multirow{2}{*}{$D$} & 
        \multicolumn{2}{c}{IGD} &
        \multicolumn{2}{c}{HV} \\ \cmidrule(lr){4-7}
        &&& NSBiDiCo & BiCo & NSBiDiCo & BiCo \\ 
    
        \midrule
        \multirow{1}{*}{DOC1}&2&6&\hl{7.2377e-3 (9.26e-4) $+$}&8.2388e-2 (2.37e-1)     & \hl{3.4323e-1 (7.80e-4) $+$}&3.0848e-1 (7.81e-2) \\
        % \hline
        \multirow{1}{*}{DOC2}&2&16&\hl{6.3262e-3 (6.47e-3) $+$}&NaN (NaN)                  & \hl{6.1782e-1 (8.76e-3) $+$}&NaN (NaN) \\
        % \hline
        \multirow{1}{*}{DOC3}&2&10&\hl{3.7413e+2 (2.22e+2) $+$}&4.9023e+2 (1.61e+2)    & \hl{3.7375e-2 (9.94e-2) $\approx$}&0.0000e+0 (0.00e+0) \\
        % \hline
        \multirow{1}{*}{DOC4}&2&8&\hl{2.1904e-2 (2.64e-3) $+$}&3.1574e-1 (4.53e-1)     & \hl{5.3774e-1 (3.25e-3) $+$}&3.2374e-1 (1.52e-1) \\
        % \hline
        \multirow{1}{*}{DOC5}&2&8&\hl{2.8400e+1 (5.50e+1) $+$}&NaN (NaN)                   & \hl{3.7903e-1 (2.04e-1) $+$}&NaN (NaN) \\
        % \hline
        \multirow{1}{*}{DOC6}&2&11&\hl{4.2941e-3 (5.69e-4) $+$}&9.5370e-1 (1.22e+0)    & \hl{5.1568e-1 (9.40e-3) $+$}&1.6457e-1 (1.94e-1)  \\
        % \hline
        \multirow{1}{*}{DOC7}&2&11&\hl{6.4501e-2 (2.47e-1) $+$}&5.5359e+0 (2.03e+0)    & \hl{4.7467e-1 (1.16e-1) $+$}&0.0000e+0 (0.00e+0)  \\
        % \hline
        \multirow{1}{*}{DOC8}&3&10&\hl{1.3517e-1 (4.59e-2) $+$}&3.1989e+1 (2.09e+1)    & \hl{7.1682e-1 (5.53e-2) $+$}&0.0000e+0 (0.00e+0)    \\
        % \hline
        \multirow{1}{*}{DOC9}&3&11&1.0723e-1 (1.02e-2) $-$&\hl{5.9704e-2 (6.25e-2)}    & NaN (NaN)&NaN (NaN)     \\
        \hline
        \multirow{1}{*} & &{$+/-/\approx$} & 8/1/0 & & 7/0/1 &  \\

        % ###################################################################################
        % ###################################################################################
        % ###################################################################################
        % ###################################################################################
        % ###################################################################################

        \hline
        \multirow{1}{*}{LIRCMOP1}&2&30&\hl{3.7855e-2 (3.41e-2) $+$}&1.6799e-1 (1.61e-2)         &   \hl{2.1722e-1 (2.15e-2) $+$}&1.5538e-1 (6.96e-3)    \\
        % \hline
        \multirow{1}{*}{LIRCMOP2}&2&30&\hl{4.1594e-2 (4.80e-2) $+$}&1.4132e-1 (1.87e-2)         &   \hl{3.3632e-1 (3.01e-2) $+$}&2.7933e-1 (1.07e-2)   \\   
        % \hline
        \multirow{1}{*}{LIRCMOP3}&2&30&\hl{8.3052e-2 (5.51e-2) $+$}&1.8971e-1 (2.88e-2)         &   \hl{1.7122e-1 (2.26e-2) $+$}&1.3404e-1 (1.03e-2)    \\
        % \hline
        \multirow{1}{*}{LIRCMOP4}&2&30&\hl{1.0327e-1 (5.10e-2) $+$}&1.7737e-1 (2.49e-2)         &   \hl{2.6991e-1 (2.35e-2) $+$}&2.3941e-1 (1.10e-2)    \\
        % \hline
        \multirow{1}{*}{LIRCMOP5}&2&30&\hl{1.0460e+0 (3.50e-1) $+$}&1.2184e+0 (6.27e-3)         &   \hl{2.6991e-1 (2.35e-2) $+$}&2.3941e-1 (1.10e-2)    \\
        % \hline
        \multirow{1}{*}{LIRCMOP6}&2&30&\hl{9.7340e-1 (4.39e-1) $\approx$}&1.3452e+0 (2.08e-4)   &   \hl{2.6525e-2 (3.12e-2) $+$}&0.0000e+0 (0.00e+0)    \\
        % \hline
        \multirow{1}{*}{LIRCMOP7}&2&30&9.2424e-1 (8.29e-1) $\approx$&\hl{5.8994e-1 (7.27e-1)}   &   1.2940e-1 (1.43e-1) $\approx$&\hl{1.7339e-1 (1.16e-1)}    \\
        % \hline
        \multirow{1}{*}{LIRCMOP8}&2&30&1.1435e+0 (7.68e-1) $-$&\hl{6.5569e-1 (6.83e-1)}         &   9.2471e-2 (1.35e-1) $\approx$&\hl{1.6005e-1 (1.07e-1)}     \\
        % \hline
        \multirow{1}{*}{LIRCMOP9}&2&30&\hl{2.5225e-1 (1.40e-1) $+$}&8.7721e-1 (1.63e-1)         &   \hl{4.5587e-1 (5.94e-2) $+$}&1.5449e-1 (8.79e-2)    \\
        % \hline
        \multirow{1}{*}{LIRCMOP10}&2&30&\hl{1.3076e-1 (2.49e-1) $+$}&8.5651e-1 (6.79e-2)        &   \hl{6.1753e-1 (1.84e-1) $+$}&9.9952e-2 (3.56e-2)    \\
        % \hline
        \multirow{1}{*}{LIRCMOP11}&2&30&\hl{2.8799e-1 (1.44e-1) $+$}&4.2997e-1 (2.13e-1)        &   \hl{5.0190e-1 (1.05e-1) $+$}&4.1979e-1 (1.44e-1)    \\
        % \hline
        \multirow{1}{*}{LIRCMOP12}&2&30&\hl{1.5634e-1 (2.24e-2) $+$}&2.7156e-1 (1.42e-1)        &   \hl{5.3916e-1 (1.53e-2) $+$}&4.9253e-1 (6.94e-2)    \\
        % \hline
        \multirow{1}{*}{LIRCMOP13}&3&30&\hl{4.7285e-1 (4.09e-1) $+$}&1.3188e+0 (1.86e-3)        &   \hl{2.7869e-1 (1.44e-1) $+$}&8.9701e-5 (9.70e-5)    \\
        % \hline
        \multirow{1}{*}{LIRCMOP14}&3&30&\hl{4.7689e-1 (4.50e-1) $+$}&1.2752e+0 (1.89e-3)        &   \hl{3.0354e-1 (1.72e-1) $+$}&4.0739e-4 (2.88e-4)    \\
        \hline
        \multirow{1}{*} & &{$+/-/\approx$} &11/1/2 & & 12/0/2 &                                  \\
        \bottomrule
    \label{table:igd_hv_results}
        
    \end{tabular}
\end{table*}

\begin{table*}[!htb]
\renewcommand{\arraystretch}{1.2}
\centering
\caption{Mean And Standard Deviation For Performance Comparisons Between NSBiDiCo And BiCo For Real-World Problems.}
\begin{tabular}{ccccc}
\toprule
Problem&$M$&$D$&NSBiDiCo&BiCo\\
\midrule
\multirow{1}{*}{Vibrating Platform Design}&2&5&\hl{3.9296e-1 (1.35e-5) $+$}&2.9783e-1 (8.17e-2)\\
% \hline
\multirow{1}{*}{Welded Beam Design}&2&4&\hl{8.6160e-1 (6.46e-4) $+$}&8.5592e-1 (6.34e-3)\\
% \hline
\multirow{1}{*}{Disc Brake Design}&2&4&4.3458e-1 (1.66e-4) $\approx$&\hl{4.3463e-1 (3.26e-4)}\\
% \hline
\multirow{1}{*}{Car Side Impact Design}&3&7&2.5926e-2 (7.73e-5) $-$&\hl{2.6108e-2 (3.48e-5)}\\
% \hline
\multirow{1}{*}{Water Resources Management}&5&3&9.4598e-2 (1.28e-3) $-$&\hl{9.6236e-2 (7.09e-4)}\\

\multirow{1}{*}{Haverly's Pooling Problem}&2&4&\hl{7.4990e-1 (1.88e-1) $+$}&NaN (NaN)\\
% \hline
\multirow{1}{*}{Reactor Network Design}&2&6&\hl{8.8148e-1 (2.73e-1) $+$}&2.8272e-1 (1.41e-1)\\
% \hline
\multirow{1}{*}{Heat Exchanger Network Design}&3&9&\hl{9.9965e-1 (1.72e-3) $+$}&4.7216e-1 (0.00e+0)\\
\hline
\multicolumn{3}{c}{$+/-/\approx$}&5/2/1&\\
\bottomrule
\end{tabular}
\label{tab:rwmops}
\end{table*}

\section{CONCLUSION}

This paper proposed a modification of BiCo, an algorithm to solve CMOPs via bidirectional coevolution. NSBiDiCo uses the differential evolution variation operator as a search engine and a new selection mechanism. Preliminary results on two benchmark test suites and eight real-world problems suggested that NSBiDiCo outperformed BiCo. Future studies will investigate more robust DE operators in order to achieve both good convergence and diversity. In addition, NSBiDiCo can be tested in more test functions and other engineering problems compared to other CMOEAs.

%%%%%%%%%%%%%%%%%%%%%%%%%%%%%%%%%%%%%%%%%%%%%%%%%%%%%%%%%%%%%%%%%%%%%%%%%%%%%%%%




% \section*{ACKNOWLEDGMENT}

% The preferred spelling of the word ÒacknowledgmentÓ in America is without an ÒeÓ after the ÒgÓ. Avoid the stilted expression, ÒOne of us (R. B. G.) thanks . . .Ó  Instead, try ÒR. B. G. thanksÓ. Put sponsor acknowledgments in the unnumbered footnote on the first page.

%%%%%%%%%%%%%%%%%%%%%%%%%%%%%%%%%%%%%%%%%%%%%%%%%%%%%%%%%%%%%%%%%%%%%%%%%%%%%%%%

\bibliographystyle{ieeetr}
\bibliography{bibliography}

\begin{thebibliography}{10}

\bibitem{liang_et_al_2023_survey}
J.~Liang, X.~Ban, K.~Yu, B.~Qu, K.~Qiao, C.~Yue, K.~Chen, and K.~C. Tan, ``A
  survey on evolutionary constrained multiobjective optimization,'' {\em IEEE
  Transactions on Evolutionary Computation}, vol.~27, no.~2, pp.~201--221,
  2023.

\bibitem{zhan2022survey}
Z.-H. Zhan, L.~Shi, K.~C. Tan, and J.~Zhang, ``A survey on evolutionary
  computation for complex continuous optimization,'' {\em Artificial
  Intelligence Review}, pp.~1--52, 2022.

\bibitem{storn_and_price_1997}
R.~Storn and K.~Price, ``Differential evolution-a simple and efficient
  heuristic for global optimization over continuous spaces,'' {\em Journal of
  Global Optimization}, vol.~11, no.~4, p.~341, 1997.

\bibitem{opara_and_arabas_2019}
K.~R. Opara and J.~Arabas, ``Differential evolution: A survey of theoretical
  analyses,'' {\em Swarm and Evolutionary Computation}, vol.~44, pp.~546--558,
  2019.

\bibitem{yu_et_al_2021}
K.~Yu, J.~Liang, B.~Qu, and C.~Yue, ``Purpose-directed two-phase multiobjective
  differential evolution for constrained multiobjective optimization,'' {\em
  Swarm and Evolutionary Computation}, vol.~60, p.~100799, 2021.

\bibitem{yang_et_al_2019_mode_sae}
Y.~Yang, J.~Liu, S.~Tan, and H.~Wang, ``A multi-objective differential
  evolutionary algorithm for constrained multi-objective optimization problems
  with low feasible ratio,'' {\em Applied Soft Computing}, vol.~80, pp.~42--56,
  2019.

\bibitem{yang_et_al_2021_mobj_de}
Y.~Yang, J.~Liu, S.~Tan, and Y.~Liu, ``A multi-objective differential evolution
  algorithm based on domination and constraint-handling switching,'' {\em
  Information Sciences}, vol.~579, pp.~796--813, 2021.

\bibitem{wang_et_al_2019_ccmode}
J.~Wang, G.~Liang, and J.~Zhang, ``Cooperative differential evolution framework
  for constrained multiobjective optimization,'' {\em IEEE Transactions on
  Cybernetics}, vol.~49, no.~6, pp.~2060--2072, 2019.

\bibitem{wang_et_al_2019_c2ode}
B.-C. Wang, H.-X. Li, J.-P. Li, and Y.~Wang, ``Composite differential evolution
  for constrained evolutionary optimization,'' {\em IEEE Transactions on
  Systems, Man, and Cybernetics: Systems}, vol.~49, no.~7, pp.~1482--1495,
  2019.

\bibitem{wamg_et_al_2011}
Y.~Wang, Z.~Cai, and Q.~Zhang, ``Differential evolution with composite trial
  vector generation strategies and control parameters,'' {\em IEEE Transactions
  on Evolutionary Computation}, vol.~15, no.~1, pp.~55--66, 2011.

\bibitem{liu_and_wang_2019}
Z.-Z. Liu and Y.~Wang, ``Handling constrained multiobjective optimization
  problems with constraints in both the decision and objective spaces,'' {\em
  IEEE Transactions on Evolutionary Computation}, vol.~23, no.~5, pp.~870--884,
  2019.

\bibitem{liu_et_al_2022}
Z.-Z. Liu, B.-C. Wang, and K.~Tang, ``Handling constrained multiobjective
  optimization problems via bidirectional coevolution,'' {\em IEEE Transactions
  on Cybernetics}, vol.~52, no.~10, pp.~10163--10176, 2022.

\bibitem{angira_2005_nsde}
R.~Angira and B.~Babu, ``Non-dominated sorting differential evolution (nsde):
  An extension of differential evolution for multi-objective optimization.,''
  in {\em IICAI}, pp.~1428--1443, 2005.

\bibitem{deb_et_al_2002}
K.~Deb, A.~Pratap, S.~Agarwal, and T.~Meyarivan, ``A fast and elitist
  multiobjective genetic algorithm: Nsga-ii,'' {\em IEEE Transactions on
  Evolutionary Computation}, vol.~6, no.~2, pp.~182--197, 2002.

\bibitem{li_and_zhang_2009}
H.~Li and Q.~Zhang, ``Multiobjective optimization problems with complicated
  pareto sets, moea/d and nsga-ii,'' {\em IEEE Transactions on Evolutionary
  Computation}, vol.~13, no.~2, pp.~284--302, 2009.

\bibitem{deb1996combined}
K.~Deb, M.~Goyal, {\em et~al.}, ``A combined genetic adaptive search (geneas)
  for engineering design,'' {\em Computer Science and informatics}, vol.~26,
  pp.~30--45, 1996.

\bibitem{kumar_2021}
A.~Kumar, G.~Wu, M.~Z. Ali, Q.~Luo, R.~Mallipeddi, P.~N. Suganthan, and S.~Das,
  ``A benchmark-suite of real-world constrained multi-objective optimization
  problems and some baseline results,'' {\em Swarm and Evolutionary
  Computation}, vol.~67, p.~100961, 2021.

\bibitem{tian_et_al_2017}
Y.~Tian, R.~Cheng, X.~Zhang, and Y.~Jin, ``Platemo: A matlab platform for
  evolutionary multi-objective optimization [educational forum],'' {\em IEEE
  Computational Intelligence Magazine}, vol.~12, no.~4, pp.~73--87, 2017.

\bibitem{fan_et_al_2020}
Z.~Fan, W.~Li, X.~Cai, H.~Li, C.~Wei, Q.~Zhang, K.~Deb, and E.~Goodman,
  ``{Difficulty Adjustable and Scalable Constrained Multiobjective Test Problem
  Toolkit},'' {\em Evolutionary Computation}, vol.~28, pp.~339--378, 09 2020.

\end{thebibliography}

\end{document}